%% file: main.tex
\documentclass{article}
\usepackage{amsmath,epsfig}
\usepackage[preprint]{spconfa4}
\usepackage{xcolor}
\usepackage{cite}

\usepackage{amsmath,amssymb,amsfonts}
\usepackage{algorithmic}
\usepackage{graphicx}
\usepackage{textcomp}
\usepackage[pagebackref,breaklinks,colorlinks,hyperfootnotes=false]{hyperref}
\usepackage{array}
\usepackage{makecell}
\usepackage{multicol}
\usepackage{multirow}
\usepackage{booktabs}
\usepackage{enumitem}

\let\OLDthebibliography\thebibliography
\renewcommand\thebibliography[1]{
  \OLDthebibliography{#1}
  \setlength{\parskip}{0pt}
  \setlength{\itemsep}{0pt plus 0.3ex}
}

\begin{document}\sloppy

% Example definitions.
% --------------------
\def\x{{\mathbf x}}
\def\L{{\cal L}}

% Title.
% ------
\title{Mix-up Self-Supervised Learning for Contrast-agnostic Applications}
%
% Single address.
% ---------------
%\name{Anonymous ICME submission}
%Address and e-mail should NOT be added in the submission paper. They should be present only in the camera ready paper. 
\name{Yichen Zhang$^{1,\ast}$, Yifang Yin$^{2,\ast}$, Ying Zhang$^{3,\dagger}$ and Roger Zimmermann$^1$}
%\name{Yichen Zhang$^{\ast}$, Yifang Yin$^{\dagger}$, Ying Zhang and Roger Zimmermann   (...) and Last Author$^{\ddagger}$}
\address{
$^1$School of Computing, National University of Singapore \\ 
$^2$Institute for Infocomm Research, A*STAR \\ 
$^3$School of Computer Science, Northwestern Polytechnical University \\
\{yichenz, dcsrz\}@nus.edu.sg, yin\_yifang@i2r.a-star.edu.sg , izhangying@nwpu.edu.cn 
%\thanks{$\ast$ Two authors contributed equally to this paper.} 
\thanks{$\ast$ Equal contribution.} 
\thanks{$\dagger$ Corresponding author.} 
}

\maketitle

\begin{abstract}
Contrastive self-supervised learning has attracted significant research attention recently. It learns effective visual representations from unlabeled data by embedding augmented views of the same image close to each other while pushing away embeddings of different images. 
Despite its great success on ImageNet classification, COCO object detection, \emph{etc.}, its performance degrades on contrast-agnostic applications, \emph{e.g.}, medical image classification, where all images are visually similar to each other. This creates difficulties in optimizing the embedding space as the distance between images is rather small. To solve this issue, we present the first mix-up self-supervised learning framework for contrast-agnostic applications. We address the low variance across images based on cross-domain mix-up and build the pretext task based on two synergistic objectives: image reconstruction and transparency prediction. Experimental results on two benchmark datasets validate the effectiveness of our method, where an improvement of 2.5\% $\sim$ 7.4\% in top-1 accuracy was obtained compared to existing self-supervised learning methods.
\end{abstract}


\section{Introduction}
\label{sec:intro}

\input{content/intro}

\section{Related Work}
\label{sec:related_work}
\input{content/related_work}

\begin{figure}[!t]
  \centering
  \includegraphics[width=.95\linewidth]{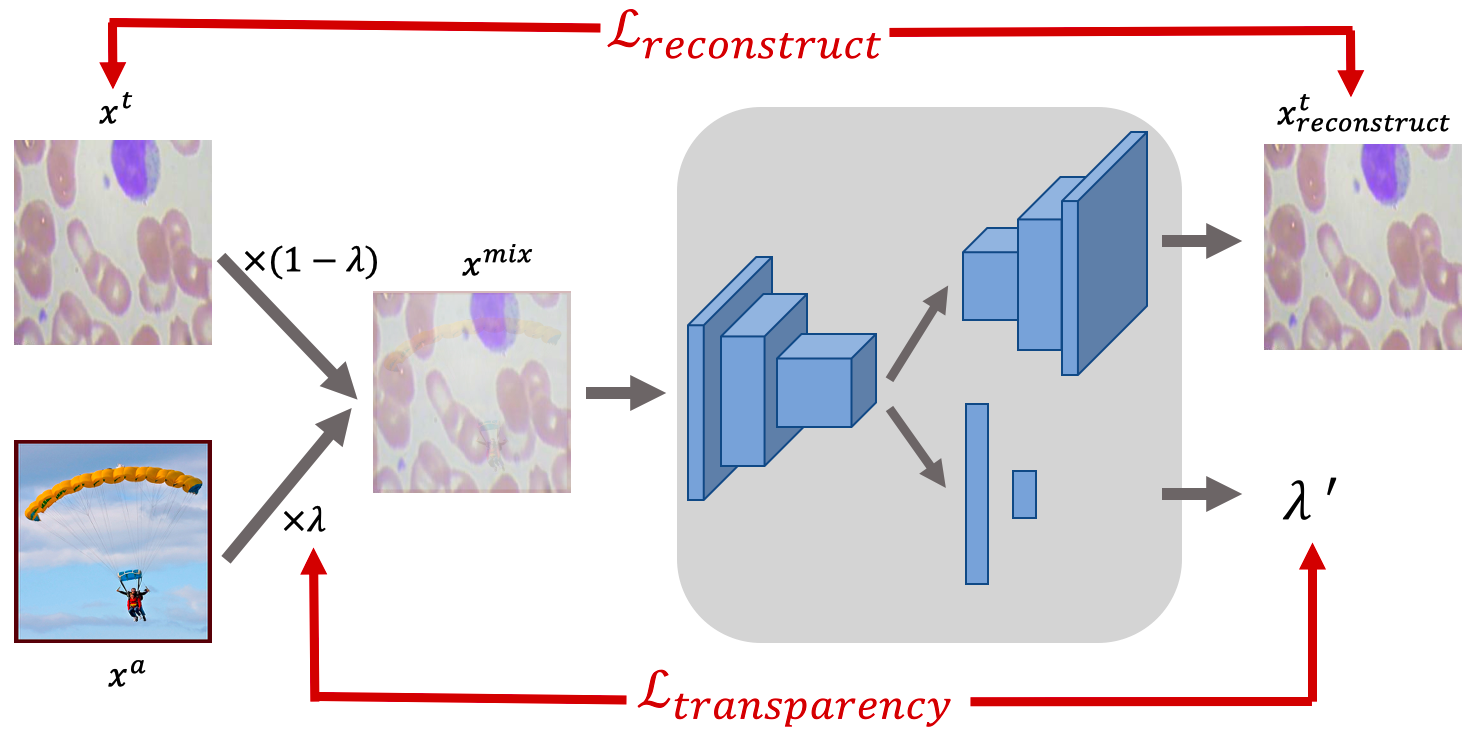}
  \caption{The proposed method for contrast-agnostic applications.}
  \label{method:fig:framework}
\end{figure}

\section{Method}
\label{sec:method}
\input{content/method}

\section{Experiments}
\label{sec:exp}

\input{content/exp}

\section{Conclusion}
\label{sec:conclusion}
\input{content/conclusion}

\section{Acknowledgement}
This research was supported by Singapore Ministry of Education Academic Research Fund Tier 1 under MOE's official grant number T1 251RES2029.

\bibliographystyle{IEEEbib}
%\footnotesize
\small
%\bibliography{macros,main}
\bibliography{main}

\appendix
\section{SUPPLEMANTARY MATERIAL}

\input{content/appendix}

\end{document}

%% file: content/intro.tex
\begin{figure}[!t]
  \centering
  \includegraphics[width=.98\linewidth]{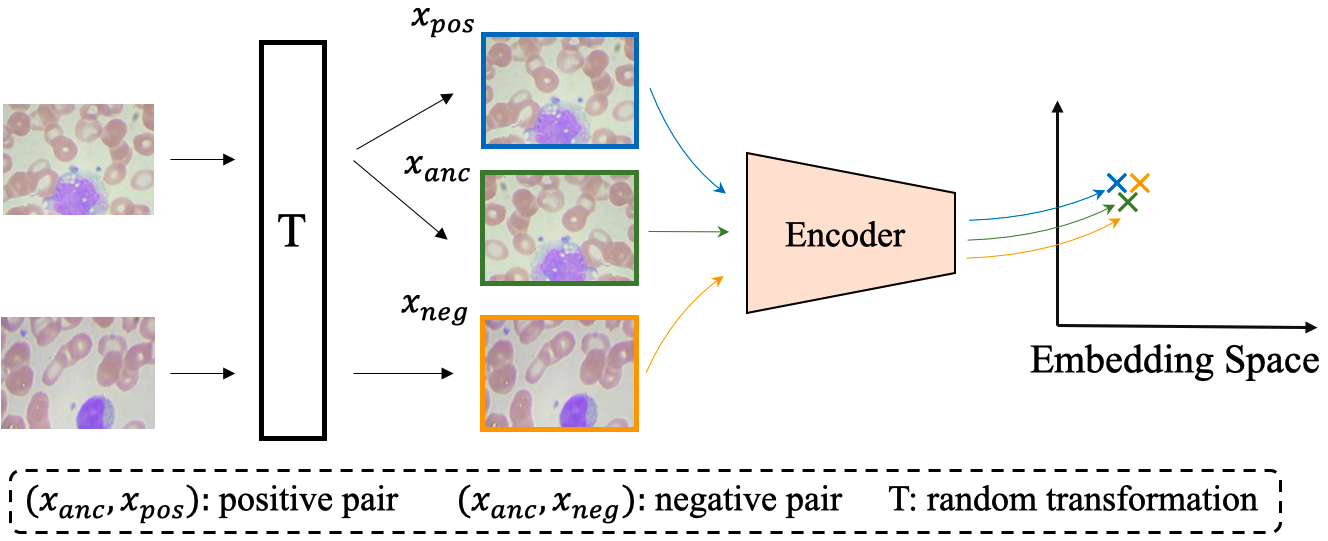}
  \caption{In contrast-agnostic applications, low variance across images can degrade the performance of contrastive self-supervised learning techniques. The high similarity between images hinders the model to distinguish the positive example $x_{pos}$ and the negative example $x_{neg}$ \emph{w.r.t.} the anchor image $x_{anc}$, resulting in obtaining less effective representations.}
  \label{intro:con-ag}
\end{figure}

Self-supervised learning has received increasing research attention from both academia and industry in recent years. It learns from unlabeled data by pretext tasks, and thus reduces the cost for building large and accurately labeled datasets~\cite{ClassificationOnNoisyLabels}.
%It also broadens the horizon of pre-trained models by utilizing multiple related datasets together, without worrying about the label inconsistency problem. In this way, knowledge and information can be leveraged and transferred across multiple domains~\cite{Feng2019SelfSupervisedRL}.

Several previous studies~\cite{Chen2020SimCLR, He2020MoCo, ACMMM2017Retargeting, Kim2020MixCoMC, dino} have achieved promising results on self-supervised visual representation learning in the image domain. State-of-the-art methods are mostly based on contrastive learning under the instance discrimination pretext task~\cite{wu2018unsupervised}. These methods aim at increasing the feature similarity between positive pairs while minimizing it between negative pairs. To achieve that, two random augmented views of the same image (\emph{i.e.}, applying random augmentation twice to the same image) are regarded as a positive pair and the augmented views of different images are regarded as negative pairs. Although good performance has been obtained on large-scale benchmark datasets such as ImageNet~\cite{deng2009imagenet} and COCO~\cite{coco}, challenges still exist in applications where all images are visually similar to each other (see Fig.~\ref{intro:con-ag}). For these applications, the high similarity among samples becomes an obstacle in optimizing the embedding space for contrastive learning as the visual gap between negative pairs is quite small. Due to this property, we refer to these applications as \emph{contrast-agnostic}. One well-known example of such applications is medical image analysis. Many of them consist of images showing certain body parts captured by specific medical instruments, resulting in visually similar images with low variance across different instances. 

Intuitively, it can be beneficial for representation learning by improving training samples' diversity in these applications. One straightforward solution is to inject random Gaussian noise into the original images. However, Gaussian noise does not contain any semantic information, resulting in minor improvements in representation learning. Instead, as a data augmentation method, mix-up~\cite{Zhang2018mixupBE} operation has been widely adopted in various applications. It generates synthetic images by mixing different images together. 
However, based on our experiments, combining mix-up with existing contrastive learning methods, e.g., mixco~\cite{Kim2020MixCoMC}, may only lead to minor improvements in contrast-agnostic applications.

To address the issues, we present \emph{MixSSL}, the first \emph{\underline{Mix}-up \underline{S}elf-\underline{S}upervised \underline{L}earning} framework for contrast-agnostic visual representation learning. 
First, we propose to enrich the diversity of the (target) training samples (\emph{e.g., medical images}) by mixing them with (auxiliary) natural images from ImageNet. This strategy not only increases the variance among training samples, but also improves the model's robustness on unseen instances. Next, we build our pretext task by fulfilling the following two synergistic objectives: image reconstruction and transparency prediction.
Reconstruction of the original training sample from the mixed input is a challenging task, which enables our model to extract the key information of the target image effectively. Meanwhile, transparency prediction predicts the mix-up ratio, which serves as an auxiliary guidance by identifying the content of the target image from the mixed input to facilitate reconstruction. Here we summarize the contributions of this paper as follows:

\begin{itemize}[leftmargin=5mm]
\item To the best of our knowledge, we are the first to propose a self-supervised learning framework for contrast-agnostic applications. Cross-domain mix-up is adopted to improve sample diversity and model robustness simultaneously.
\item We build our pretext task by jointly performing image reconstruction and transparency prediction. Thus, our model learns to distinguish visual content from different domains and further extract key information from the target domain.
\item We perform comprehensive experiments on two contrast-agnostic image datasets. The experimental results show that our method outperforms the state-of-the-art self-supervised learning methods by a large margin on both datasets.
\end{itemize}

%% file: content/related_work.tex
Since good representations are the foundation of good results~\cite{PAMI_Representation,yin2021enhanced}, a significant number of self-supervised representation learning techniques have been proposed by researchers world-wide in the image domain. Here we roughly divide them into two categories: non-contrastive-based and contrastive-based methods, depending on whether contrastive learning has been adopted for building pretext tasks. 

\textbf{Non-contrastive-based methods.} 
To circumvent the requirement of the labeling process, many pretext tasks have been proposed to train the model in an unsupervised manner.
Most of them were inspired by image-level operations and the internal property of data, which provide temporary target for the model training.
For example, researchers constructed the training target via predicting the rotation angle~\cite{Gidaris2018UnsupervisedRL} or solving the Zigzaw problem~\cite{Noroozi2016UnsupervisedLO}, whose answer can either be derived from the augmentation process or the original image.
Also, other examples include context prediction~\cite{Doersch2015UnsupervisedVR}, affine transformation~\cite{Novotn2018SelfSupervisedLO} and colorization~\cite{Zhang2016ColorfulIC}.

\textbf{Contrastive-based methods.} 
As one of the most important branches in self-supervised learning, many studies based on contrastive learning have been developed under the instance discrimination pretext task (matching the encoded views of the same image). They mainly focused on finding a better way to construct or utilize positive or negative pairs for model training.
SimCLR~\cite{Chen2020SimCLR} adopted the non-linear projection head between the representation and the loss function. MoCo~\cite{He2020MoCo} maintained a key encoder and a corresponding large memory bank (dictionary) to improve stability. Dino~\cite{dino} adopted a teacher network to facilitate model training. MixCo~\cite{Kim2020MixCoMC} extended the contrastive learning concept to semi-positives encoded from the mix-up~\cite{Zhang2018mixupBE} of positive and negative images.
Although these methods achieve state-of-the-art results on the ImageNet dataset, they perform unsatisfactorily on contrast-agnostic applications.

%% file: content/method.tex
\subsection{Overview}
Self-supervised learning, especially contrastive self-supervised learning, encounters serious challenges when applied to contrast-agnostic applications, due to the small visual gap among samples.  
In this section, we introduce \emph{\underline{Mix}-up \underline{S}elf-\underline{S}upervised \underline{L}earning} (\emph{MixSSL}) framework for contrast-agnostic applications. As illustrated in Fig.~\ref{method:fig:framework}, our method deals with the low-variance challenge by first mixing samples from different domains and then extracting key information of the target domain based on two synergistic objectives. 

Formally, given a target contrast-agnostic dataset $X^t=\{x_1^t, x_2^t, \dots, x_n^t \}$ and an auxiliary dataset $X^a=\{x_1^a, x_2^a, \dots, x_m^a \}$, our goal is to learn good visual representations for the target dataset $X^t$ with the facilitation of the auxiliary dataset $X^a$. Since our method is a self-supervised visual representation learning approach, labels from both datasets are not required.

\subsection{MixSSL}
Our overall objective is to obtain a well-trained encoder that performs outstandingly at extracting informative visual representations for a target contrast-agnostic application. 
To achieve this goal, we design our pretext task as a composition of two correlated objectives with cross-domain mix-up applied in data preprocessing. The main objective is the image reconstruction task, which aims to extract the key semantic information from target domain images. The auxiliary objective is the transparency prediction task, which aims to recognize the composition ratio from the mixed samples so as to facilitate target image reconstruction.
The overall framework is illustrated in Fig.~\ref{method:fig:framework}, which consists of a shared encoder for feature extraction, a decoder for image reconstruction and a linear regressor for transparency prediction.

\textbf{Cross-domain mix-up.}
The mix-up operation was first proposed in \cite{Zhang2018mixupBE} as an image augmentation method.
It simultaneously mixes the training samples and their labels as,
\begin{align*}
    x &= \lambda x_i + (1-\lambda) x_j \\
    y &= \lambda y_i + (1-\lambda) y_j
\end{align*}
where $\lambda$ is sampled from the Beta distribution. $x$ and $y$ are samples and labels, respectively. The training of a model with this strategy is capable of reducing the amount of undesirable oscillation when meeting with unseen training samples~\cite{Zhang2018mixupBE}.
We build upon the conventional mix-up strategy and present a cross-domain variant to improve the diversity of our training samples. Our cross-domain mix-up method mixes samples from different domains:
\begin{equation}
\label{eqn:mix}
    x^{mix} =  (1-\lambda) x^t + \lambda x^a 
\end{equation}
where $x^t$ and $x^a$ denote two samples from the target dataset and the auxiliary dataset, respectively. The mix-up ratio $\lambda$ is again sampled from the Beta distribution, varying from batch to batch. Different from \cite{luo2020generalizing}, where cross-domain mix-up serves as a solution to the domain shift, our cross-domain mix-up method effectively increases the variance among samples in contrast-agnostic applications, but may distract the model from learning specific distribution of the target domain. 
To solve this issue, we build our pretext task by jointly performing image reconstruction and transparency prediction. Thus, our model learns to separate the content of $x^t$ and $x^a$ from the mixed input $x^{mix}$ and capture the key semantic information of $x^t$ at the same time.

\textbf{Image reconstruction.}
As aforementioned, existing contrastive learning methods may confuse the model in contrast-agnostic applications. We alternatively adopt image reconstruction as our main pretext task for self-supervised learning. The reconstruction loss is defined as the distance between the original target image $x^t$ and the reconstructed target image $x^t_{reconstruct}$ from the mixed input $x^{mix}$ as,
\begin{equation}
    \L_{reconstruct}=\mathcal{D}(Decoder(Encoder(x^{mix})), x^t)
\end{equation}
where $\mathcal{D}$ is a distance metric measuring the difference between original and reconstructed images, which is normally implemented as a mean square error (MSE) function. 
It is worth noting that our proposed method is different from the traditional autoencoder and decoder as we reconstruct an image from its faded version obtained by cross-domain mix-up. The advantages of our method are twofold. First, it motivates the model to separate the mixed input by learning to select the content that is originally from the target image. Second, the key semantic information of target images are extracted by the encoder to perform reconstruction. Facilitated by the transparency prediction task to be introduced in the next section, our model learns more effective visual representations compared to existing contrastive learning based methods.

\textbf{Transparency prediction.}
Recall that we construct training samples by mixing images from different domains as shown in Fig.~\ref{method:fig:framework}. The mix-up operation combines the target and the auxiliary images by linear interpolation with weight $\lambda$. We subsequently introduce an auxiliary pretext task, termed transparency prediction, to predict the weight $\lambda$ based on the mixed input sample. It is beneficial to jointly optimize image reconstruction and transparency prediction as the two tasks are intrinsically correlated with each other. Here we model the transparency prediction task as a regression problem and predict $\lambda$ based on multi-layer perception (MLP). The loss function is given by,
\begin{align}
    &\lambda' = MLP(Encoder(x^{mix})) \\
    &\L_{transparency}=|\lambda'- \lambda|
\end{align}
where $|\lambda'-\lambda|$ is the absolute difference between the predicted transparency $\lambda'$ and the ground-truth transparency $\lambda$.

\textbf{Overall loss function.}
In summary, the above two objectives encourage the encoder to extract well-trained visual representations from the target images based on self-supervised learning. The overall loss is defined as a linear combination of $\L_{reconstruct}$ and $\L_{transparency}$ as:
\begin{equation}
    \L_{pretrain}=\L_{reconstruct}+\gamma\L_{transparency}
\end{equation}
where $\gamma$ is a hyper-parameter balancing the impact of the two losses.

%% file: content/exp.tex
\begin{figure}[!t]
  \centering
  \includegraphics[width=0.65\linewidth]{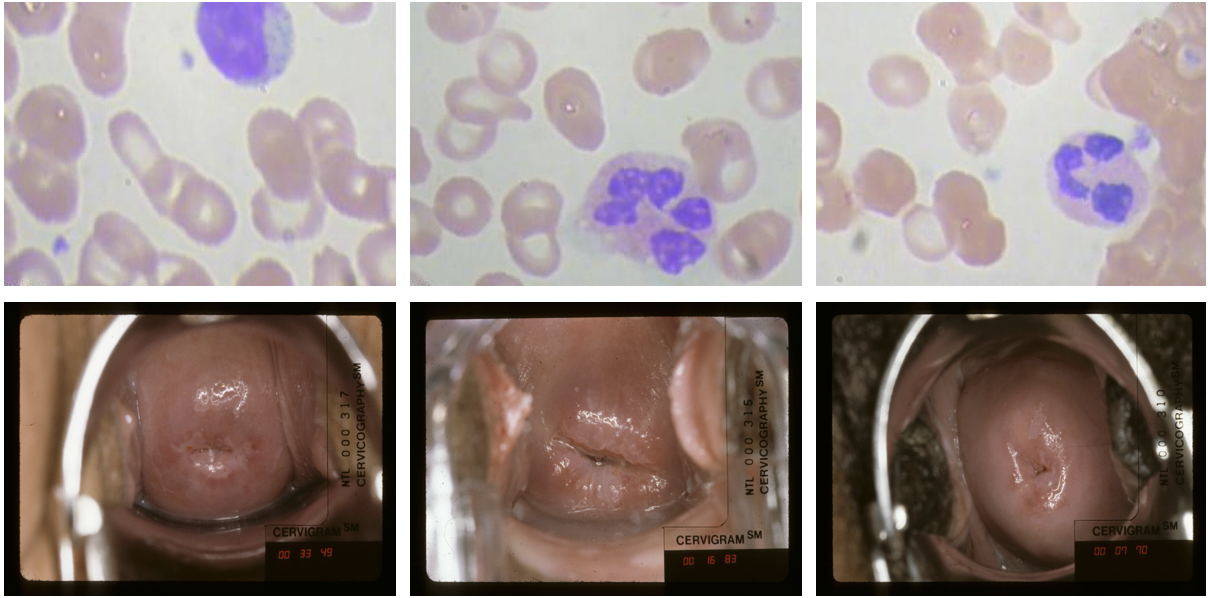}
  \caption{Data samples from two datasets, including the blood cell dataset (first row) and the cervix dataset (second row).}
  \label{exp::fig::sample}
\end{figure}

\input{tab/main_bloodcell}

\subsection{Experimental setup}
We adopt two benchmark datasets to evaluate our proposed mix-up self-supervised learning framework, including the blood cell dataset and the cervix dataset. They can both be regarded as contrast-agnostic applications due to the high visual similarity across samples. Examples from the two datasets are shown in Fig.~\ref{exp::fig::sample}. 

\textbf{Blood cell dataset}~\cite{bloodcell-dataset} is a public dataset consisting of four types of blood cell images. 
It contains nearly 10,000 images for training and nearly 2,500 images for validation. We train our model on the training set and report the results on the validation set.

\textbf{Cervix dataset} consists of four related datasets: NHS~\cite{Herrero2000PopulationbasedSO}, ALTS~\cite{Walker2003ART}, CVT~\cite{cvt} and Biopsy~\cite{biopsy}. 
A total of nearly 20,000 cervical images are included. Following previous work~\cite{zhang2021spatial, zhang2022evaluation}, we split out a validation set from the NHS dataset for evaluation and assign binary labels to them according to their original labels.
More details about the cervix data description can be found in supplementary material.

\input{tab/main_cervix}

We compare our framework with five state-of-the-art self-supervised learning methods~\cite{Gidaris2018UnsupervisedRL, Chen2020SimCLR, He2020MoCo, dino, Kim2020MixCoMC} on two datasets.
%a) \textbf{Rotation}~\cite{Gidaris2018UnsupervisedRL} proposed the rotation prediction pretext task.
%b) \textbf{SimCLR}~\cite{Chen2020SimCLR} adopted non-linear projection heads for contrastive learning. 
%c) \textbf{MoCo}~\cite{He2020MoCo} proposed the momentum encoder structure and the usage of the dictionary. 
%d) \textbf{Dino}~\cite{dino} adopted a teacher network to facilitate training.
%e) \textbf{MixCo}~\cite{Kim2020MixCoMC} increased the task difficulty by combining the mix-up strategy with the MoCo structure.

\subsection{Implementation details}
Following previous work~\cite{He2020MoCo}, we adopt ResNet-50~\cite{He2016resnet} as our encoder and ResNet-Generator from CycleGAN~\cite{CycleGAN2017} as our decoder. The linear regressor contains two linear layers with 128 and 1 units, respectively.
All methods including ours are first initialized with an ImageNet pre-trained model Dino~\cite{dino}. Next, they are trained to learn visual representations on the contrast-agnostic datasets based on pretext tasks, and finally fine-tuned for the downstream classification task with class labels.
The parameter $\gamma$ in the loss function is empirically set to 1 throughout the experiments. % as we found it beneficial to balance two objectives.
For the auxiliary dataset to perform cross-domain mix-up, we adopt ImageNette~\cite{imagenette}, which is a subset of the ImageNet consisting of 10 classes only.
For more details about data preprocessing and training please refer to supplementary material.

\subsection{Comparison to the state-of-the-art}
We compare our framework with five state-of-the-art self-supervised learning methods on both the blood cell dataset and the cervix dataset, and report the results in Tables~\ref{exp:table:bloodcell} and~\ref{exp:table:cervix}. We also report the results obtained by using random initialization (Rand) or self-supervised pre-trained weights on ImageNet (ImageNet-self) as baselines for comparison.

Table~\ref{exp:table:bloodcell} reports both the per-class and average top-1 accuracy, together with the average performance gain against the baseline Rand on the blood cell dataset.
We observe that our method outperforms all the other methods in terms of the average accuracy. For the most difficult class, \emph{i.e.}, the Monocyte, our method obtains a promising result of 85.81\%, while its competitors only achieves top-1 accuracy of less than 80\%.
Compared to the second best solution~\cite{Kim2020MixCoMC}, our method outperforms it by 1.5\% in Eosinophil, 7.91\% in Monocyte, 0.64\% in Neutrophil, and 2.53\% in average accuracy.

Table~\ref{exp:table:cervix} shows the result comparison on the cervix dataset. In addition to the top-1 accuracy, we also report the precision, recall, F1 score, and ROC-AUC since the cervix cancer detection is a binary classification problem. ROC-AUC is an essential metric for medical applications, which reveals the overall model capability. As can be seen, our method obtains the best result in terms of F1 score, ROC-AUC, and accuracy. Compared to the second best solution~\cite{Kim2020MixCoMC}, our model achieves an improvement of 9.49\% in recall, 4.40\% in F1 score, 2.14\% in ROC-AUC, and 2.84\% in accuracy.

Additionally, we observe that self-supervised models trained on the ImageNet (ImageNet-self) outperforms most of the self-supervised models trained on medical images. One possible reason is that existing contrastive-based approaches suffer from the low-variance challenge among medical images, which makes them less effective in visual representation learning for such contrast-agnostic applications. 
Our method, on the other hand, addresses the low-variance challenge by applying cross-domain mix-up assisted with two complementary objectives, leading to a better performance when being fine-tuned for downstream tasks.

\subsection{Ablation studies}
In this section, we conduct ablation studies to investigate the usage of mix-up operation and the design of pretext task.

\input{tab/abla_obj}

\textbf{Mix-up operation.} 
Recall that we apply mix-up to make training samples more diverse in our framework. Then a question naturally arises: can mix-up also improve the performance of existing contrastive self-supervised methods? To find the answer, we compare SimCLR and MoCo with their mix-up variants Mixup+SimCLR and MixCo in Tables~\ref{exp:table:bloodcell} and~\ref{exp:table:cervix}.
The results show that the mix-up operation only leads to minor improvements in the SimCLR and MoCo frameworks. Our proposed method outperforms Mixup+SimCLR and MixCo by 2.48\% $\sim$ 6.25\% in terms of the top-1 accuracy, which indicates that our method leverages the mix-up operation to deal with the challenges in a more effective way.
Reconstructing the original images from the mixed input where only partial information is accessible is a challenging task. By training based on this objective, our encoder has stronger generalization capability and is able to extract informative representations of target images better compared to other methods.

\textbf{Pretext task.}
In addition to the image reconstruction ($\mathcal{R}$) and the transparency prediction ($\mathcal{T}$) adopted in our framework, we also evaluate the effectiveness of two alternatives tasks, namely contrastive learning ($\mathcal{C}$) and auxiliary label prediction ($\mathcal{A}$). Results are reported in Table~\ref{exp:table:abla_obj}. As it shows, both $\mathcal{R}$ and $\mathcal{T}$ are essential in our proposed framework, missing either of them results in performance degradation of more than 2\% in terms of accuracy. The experimental results also indicate that the image reconstruction task $\mathcal{R}$ is more important for visual representation learning of medical images. The transparency prediction task $\mathcal{T}$ can be considered as an auxiliary objective to facilitate the reconstruction as both of them are intrinsically correlated with each other.

Next, we compare our transparency prediction task $\mathcal{T}$ with its alternatives $\mathcal{C}$ and $\mathcal{A}$. $\mathcal{C}$ utilizes the contrastive loss that narrows the gap between two augmented views, while $\mathcal{A}$ leverages auxiliary label information as an instruction.
By jointly optimizing with the reconstruction loss, our transparency prediction task $\mathcal{T}$ outperforms $\mathcal{C}$ and $\mathcal{A}$ by 2.97\% and 1.80\% on the blood cell dataset and by 5.68\% and 3.97\% on the cervix dataset. Task $\mathcal{C}$ optimizes the feature space based on mixed samples without distinguishing the content from different domains. Task $A$ is a supervised task where labels of the auxiliary dataset are required. Our proposed transparency prediction task $\mathcal{T}$ addresses the drawbacks of $\mathcal{C}$ and $\mathcal{A}$, which has been shown to be more effective in visual representation learning for contrast-agnostic applications.

%% file: tab/main_bloodcell.tex
\begin{table*}[!t]
	\centering
	%>{\centering\arraybackslash}m{1.2in}
	%\begin{tabular}{|c|>{\centering\arraybackslash}m{0.65in}|c|c|c|c|c|}
	%\begin{tabular}{|c|c|c|c|c|c|c|}
	\begin{tabular}{|>{\centering\arraybackslash}m{1.1in}|c|>{\centering\arraybackslash}m{0.6in}|>{\centering\arraybackslash}m{0.7in}|>{\centering\arraybackslash}m{0.6in}|>{\centering\arraybackslash}m{0.6in}|>{\centering\arraybackslash}m{1.1in}|}
	%\Xhline{1.1pt}
	\hline
		%\multirow{2}{*}{Pre-trained model} & \multirowcell{2}{Linear\\Evaluation} & \multicolumn{5}{c|}{Top-1 Accuracy (\%)} \\
		%\cline{3-7}
		%&& Eosinophil &  Lymphocyte & Monocyte & Neutrophil & Average  \\
		\multirow{2}{*}{Pre-trained model} & \multirow{2}{*}{Mix} & \multicolumn{5}{c|}{Top-1 Accuracy (\%)} \\
		\cline{3-7}
		&& Eosinophil &  Lymphocyte & Monocyte & Neutrophil & Average  \\
		
		%\hline\hline
		%\Xhline{0.8pt}
		\hline
		Rand &$\times$&76.40&99.19&75.02&94.55& 86.29 (-)  \\
		SimCLR~\cite{Chen2020SimCLR}  &$\times$&80.26&99.84&83.39&91.99& 88.86 (+1.94\%)  \\
		%Mixup+SimCLR & 88.28 & 100.00 & 75.02 & 93.59 & 89.22 (+2.93\%) \\
		%MoCo &  &&& \\
		%Mixup + MoCo & &&& \\
        Dino~\cite{dino} &$\times$&87.60&100.00&74.68&94.39&89.22 (+2.93\%)  \\
        MoCo~\cite{He2020MoCo} &$\times$& 83.79 & 100.00 & 77.10 & \textbf{96.15} & 89.26 (+2.97\%)  \\
        Rotation~\cite{Gidaris2018UnsupervisedRL} &$\times$&87.48 & 100.00 & 75.48 & 94.07 & 89.26 (+2.97\%)  \\
        %Noise+SimCLR  &$\times$&87.80&100.00&75.02&94.39& 89.30 (+3.01\%)  \\
        ImageNet-self &$\times$&85.07&100.00&78.71&93.91 & 89.43 (+3.14\%)  \\
        \hline
        Mixup+SimCLR &$\surd$& \textbf{88.28} & 100.00 & 75.02 & 93.59 & 89.22 (+2.93\%) \\
        MixCo~\cite{Kim2020MixCoMC}&$\surd$&86.19&100.00&77.90&94.71& 89.71 (+3.42\%)  \\
        \hline
		%Ours w/o $\L_{reconstruct}$ &$\surd$& \textbf{88.28} &100.00 & 75.16 & 93.75 & 89.26 (+2.97\%) \\
		%Ours w/o $\L_{transparency}$ &$\surd$& 86.51 & 99.84 & 75.00 & \textbf{97.12} & 89.63 (+3.34\%) \\
		\textbf{MixSSL (Ours)} &$\surd$&87.69&\textbf{100.00}&\textbf{85.81}&95.35 & \textbf{92.19} (+5.90\%)  \\
		\hline%\Xhline{1.1pt}
	\end{tabular}
	\caption{Performance comparison between our framework and state-of-the-art methods on the blood cell dataset.}
	
	\label{exp:table:bloodcell}
\end{table*}

%% file: tab/main_cervix.tex
\begin{table*}[!t]
	\centering
	%>{\centering\arraybackslash}m{1.2in}
	\begin{tabular}{|>{\centering\arraybackslash}m{1.1in}|c|>{\centering\arraybackslash}m{0.6in}|>{\centering\arraybackslash}m{0.6in}|>{\centering\arraybackslash}m{0.6in}|>{\centering\arraybackslash}m{0.7in}|>{\centering\arraybackslash}m{1.1in}|}
	%\Xhline{1.1pt}
	\hline
    	\multirow{2}{*}{Pre-trained model} & \multirow{2}{*}{Mix} & \multicolumn{5}{c|}{Metrics (\%)} \\
		\cline{3-7}
		%Pre-trained model &  Precision & Recall & F1 score & ROC-AUC & Accuracy  \\
		&&  Precision & Recall & F1 score & ROC-AUC & Accuracy  \\
		%\hline\hline
		%\Xhline{0.8pt}
		\hline
		Rand &$\times$&63.32&68.51&65.88&75.70&70.45 (-)  \\
		SimCLR~\cite{Chen2020SimCLR}&$\times$ & 71.82& 83.69& 77.21& 82.26& 79.55 (+9.10)\\
		%Noise+SimCLR &$\times$ &71.62&86.39&78.39&83.91 &80.11 (+9.66)\\
		%Mixup+SimCLR &$\surd$&80.68& 75.31 & 79.55& 77.36& 83.91        \\
		%MoCo &  &&& \\
		%Mixup + MoCo & &&& \\
        Dino~\cite{dino} &$\times$&75.31& 87.72& 81.06& 84.47 &82.95 (+12.50) \\
        MoCo~\cite{He2020MoCo} &$\times$& 73.12& \textbf{93.26}& 81.91& 85.90 &82.95 (+12.50) \\
        Rotation~\cite{Gidaris2018UnsupervisedRL}&$\times$& 81.46&78.17& 79.71&86.12  &83.52 (+13.07)\\
        ImageNet-self &$\times$&79.72&80.86&80.35&86.21 &83.52 (+13.07)\\
        \hline
        Mixup+SimCLR &\checkmark& 75.31 & 79.55& 77.36& 83.91&80.68 (+10.23)  \\
        MixCo~\cite{Kim2020MixCoMC}&\checkmark&\textbf{81.72}&79.53&80.61&86.92 &84.09 (+13.64)\\
        \hline
		%Ours  & \textbf{90.993} (+2.654\%) && \\
		%Ours w/o $\L_{reconstruct}$ &$\surd$&\textbf{82.13}&75.39&78.61&87.26& 82.95 (+12.50)\\
		%Ours w/o $\L_{transparency}$ &$\surd$&76.52&89.05&82.31&85.59& 84.09 (+13.64)\\
		\textbf{MixSSL (Ours)} &\checkmark&81.27&89.02&\textbf{85.01}&\textbf{89.06} &\textbf{86.93} (+16.48)\\
		\hline%\Xhline{1.1pt}
	\end{tabular}
	\caption{Performance comparison between our framework and state-of-the-art methods on the cervix dataset.}
	
	\label{exp:table:cervix}
\end{table*}

%% file: tab/abla_obj.tex
\begin{table}[!t]
	\centering
	%>{\centering\arraybackslash}m{1.2in}
	%\begin{tabular}{|c|>{\centering\arraybackslash}m{0.65in}|c|c|c|c|c|}
	\begin{tabular}{c|cccc|c}
	%\Xhline{1.1pt}
	%\hline
	\toprule
		%\multirow{2}{*}{Pre-trained model} & \multirowcell{2}{Linear\\Evaluation} & \multicolumn{5}{c|}{Top-1 Accuracy (\%)} \\
		%\cline{3-7}
		%&& Eosinophil &  Lymphocyte & Monocyte & Neutrophil & Average  \\
		%\multirow{2}{*}{Dataset} & \multicolumn{4}{c|}{Objective} & \multirow{2}{*}{Accuracy (\%)} \\
		%\cline{2-5}
		Dataset & $\mathcal{R}$ &  $\mathcal{T}$ & $\mathcal{C}$ & $\mathcal{A}$ & Accuracy (\%)  \\
		
		%\hline\hline
		%\Xhline{0.8pt}
		\midrule
		%\hline
		\multirowcell{5}{Bloodcell\\dataset}
		&&\checkmark&&&89.26  \\
		&\checkmark&&&&89.63  \\
		&\checkmark&\checkmark&&&\textbf{92.19} \\
		%\cline{2-5}
		&\checkmark&&\checkmark&&89.22 \\
		&\checkmark&&&\checkmark&90.39 \\
		\midrule
		%\hline
		%\hline
		\multirowcell{5}{Cervix\\dataset}
		&&\checkmark&&&82.95  \\
		&\checkmark&&&&84.09  \\
		&\checkmark&\checkmark&&&\textbf{86.93} \\
		%\cline{2-5}
		&\checkmark&&\checkmark&&81.25 \\
		&\checkmark&&&\checkmark&82.96 \\
        
		%\hline%\Xhline{1.1pt}
		\bottomrule
	\end{tabular}
	\caption{Performance comparison between different objectives. $\mathcal{R}$: image reconstruction; $\mathcal{T}$: transparency prediction; $\mathcal{C}$: contrastive learning; $\mathcal{A}$: auxiliary label prediction.}
	
	\label{exp:table:abla_obj}
\end{table}

%% file: content/conclusion.tex
In this paper, we propose a novel self-supervised visual representation learning framework for contrast-agnostic applications. Our method deals with the low-variance challenge by applying cross-domain mix-up to diversify training samples. Furthermore, we build our pretext task based on image reconstruction and transparency prediction to learn representations in the target domain. 
Extensive experiments have been conducted on two benchmark datasets. Results show that our method outperforms existing contrastive self-supervised learning methods by a large margin.

%% file: content/appendix.tex
\subsection{Cervix dataset}
\label{app:cervix_dataset}

We collect images from NHS~\cite{Herrero2000PopulationbasedSO}, ALTS~\cite{Walker2003ART}, CVT~\cite{cvt} and Biopsy~\cite{biopsy} as a cervix dataset for pre-training. They were collected from four separate medical studies by the National Cancer Institute (NCI) during previous decades. 
\begin{itemize}
    \item Natural History Study of HPV and Cervical Neoplasia (NHS) is a longitudinal study in Costa Rica started in June 1993, which focuses on studying the role of human papillomavirus infection in the etiology of high grade cervical neoplasia and to evaluate new cervical cancer screening technologies. During 7 years, 10,000 women were enrolled. It ends up consisting of nearly 1,000 general cervical images with the dimensions of around $2400\times1600\times3$. 
    \item ASCUS-LSIL Triage Study (ALTS) was designed to evaluate 3 alternative methods (immediate colposcopy, repeat PAP tests and testing for HPV) for managing atypical squamous cells of undetermined significance (ASCUS) and low grade squamous intraepithelial lesions (LSIL). It is a randomized clinical trial started in November 1996, where women age 18+ with ASCUS ($n=3,488$) or LSIL ($n=1,572$) cytology were enrolled at 4 colposcopy clinics in the United States.
    \item Costa Rica Vaccine Trial (CVT) was a double-blind study conducted to assess the efficacy of the HPV vaccine for cervical cancer prevention in women 18-25 years old. Totally 7,466 women were enrolled in the study since January 2010 and were randomized to receive the HPV 16/18 vaccine or the Hepatitis A vaccine at 0, 1 and 6 months.
    \item Biopsy Study (Biopsy) was a cross-sectional study designed to understand cervical disease on the lesion level and to establish criteria for conducting cervical biopsies. Out of 2,270 women referred for colposcopy, 690 eligible women consented to participate in the study.
\end{itemize}
During these projects, each patient may have participated in multiple screening sessions, where two photographs of the cervix (cervigrams) were taken during each recruitment and clinic visit. 
Totally nearly 20,000 cervical images are included. 
The cervical intraepithelial neoplasia (CIN) level normally serves as the criterion to judge the severity of cervical cancer. In our dataset, cervical images are labeled from CIN0 to CIN4, where histologic CIN2 or worse (CIN2+: CIN2, CIN3, CIN4) indicates the cancer precursor. To construct an appropriate dataset for the model training, which aims at alerting potential patients for further medical examination, we model this problem as a binary classification problem. Cases with CIN2+ are regarded as positive cases, while others are regarded as negative cases. Also, positive cases whose screening date surpass one year are discarded due to the possible noise induced by these samples. In this way, we can obtain 885 images for NHS dataset and 15,724 images for ALTS dataset. A train-test ratio of 4:1 is further employed on these datasets following \cite{zhang2021spatial}. 
The accessibility of these datasets is based on request and constrained agreement.

\subsection{Implementation detail}
\label{app:training}
\textbf{Data preprocessing.}
We adopt ImageNette~\cite{imagenette}, which is a subset of the ImageNet dataset consisting of only 10 easily classified classes.
During preprocessing, due to the high resolution of cervical images of $2,400\times1,600$, we train a cervix detector following \cite{ALYAFEAI2020autopipeline} and use the detector to focus on the important area. 
All the input images are put through some basic data augmentation methods, including resize and crop to $224\times224$ randomly and random horizontal flip. For the self-supervised pre-training, we also adopt some extra augmentation steps following previous works, including random color jitter, random gray scale and gaussian blur. 
The cervical images are also processed by the extra augmentation steps empirically due to the superior performance we obtain by applying these methods.

\textbf{Training.}
All of the self-supervised learning methods in our experiments, including our framework, are initialized with the ImageNet self-supervised pre-trained model from Dino~\cite{dino}, and then sent through a pre-training process for 20 epochs to inject domain-specific knowledge before starting the fine-tuning process. $\alpha$ and $\gamma$ in the loss function are both empirically set to 1 during our experiments. 
We adopt an initial learning rate of $1^{-4}$ with weight decay set to $1^{-4}$ using Adam optimizer. The batch size is set to 64 and epoch is set to 20 for both pre-training and fine-tuning.